\documentclass[11pt]{article}

\usepackage[preprint]{acl}

\usepackage{times}
\usepackage{latexsym}

\usepackage[T1]{fontenc}
\usepackage[utf8]{inputenc}

\usepackage{microtype}

\usepackage{inconsolata}

\usepackage{graphicx}
\usepackage{amsmath}
\usepackage{amsthm}
\usepackage{amssymb}
\usepackage{booktabs}
\usepackage{pifont}
\usepackage{color}
\usepackage{tcolorbox}
\usepackage{tabularx}
\usepackage{soul}
\usepackage{xcolor}
\usepackage{multirow}
\tcbuselibrary{breakable}
\tcbuselibrary{skins}
\usepackage[table]{xcolor}
\usepackage{subfigure}

\newtheorem{definition}{Definition}
\title{The Gray Zone of Faithfulness:\\Taming Ambiguity in Unfaithfulness Detection}

\author{
  \textbf{Qiang Ding},
  \textbf{Lvzhou Luo},
  \textbf{Yixuan Cao\textsuperscript{*}},
  \textbf{Ping Luo\textsuperscript{*}}
\\
  Key Lab of Intelligent Information Processing, Institute of Computing Technology,\\Chinese Academy of Sciences (CAS), Beijing 100190, China
\\
  State Key Lab of Al Safety, Beijing 100094, China
\\
  University of Chinese Academy of Sciences, Beijing 100049, China
\\
 \texttt{\{dingqiang22z,luolvzhou23s,caoyixuan,luop\}@ict.ac.cn}
}

\begin{document}
\maketitle
\begin{abstract}
Ensuring that Large Language Models (LLMs) generate summaries faithful to a given source document is essential for real-world applications. 
While prior research has explored LLM faithfulness, existing benchmarks suffer from annotation ambiguity, primarily due to the ill-defined boundary of permissible external knowledge in generated outputs.
For instance, common sense is often incorporated into responses and labeled as ``faithful'', yet the acceptable extent of such knowledge remains unspecified, leading to inconsistent annotations.
To address this issue, we propose a novel faithfulness annotation framework, which introduces an intermediate category, \texttt{Out-Dependent}, to classify cases where external knowledge is required for verification.
Using this framework, we construct \textbf{VeriGray}\footnote{\url{https://huggingface.co/datasets/Ding-Qiang/veri-gray}} (\textbf{Veri}fication with the \textbf{Gray} Zone) -- a new unfaithfulness detection benchmark in summarization.
Statistics reveal that even SOTA LLMs, such as GPT-5, exhibit hallucinations ($\sim 6\%$ of sentences) in summarization tasks.
Moreover, a substantial proportion ($\sim 9\%$ on average of models) of generated sentences fall into the \texttt{Out-Dependent} category, underscoring the importance of resolving annotation ambiguity in unfaithfulness detection benchmarks.
Experiments demonstrate that our benchmark poses significant challenges to multiple baseline methods, indicating considerable room for future improvement.
\end{abstract}

\section{Introduction}

Knowledge-grounded generation, such as summarization, extends LLMs' application to domain-specific tasks, where faithfulness to the provided source knowledge is critical, especially in high-stakes fields like finance and healthcare.
However, SOTA LLMs can still generate content unfaithful to given knowledge sources, which yields the faithfulness hallucination (or unfaithfulness for brevity) detection task \citep{niu2024ragtruth, cossio2025comprehensive}.

Unlike general hallucination detection, \textit{unfaithfulness detection} specifically identifies errors unfaithful to a given knowledge source, e.g., a document.
To evaluate unfaithfulness detectors, various benchmarks have been proposed for tasks including knowledge question answering \citep{dziri2022evaluating, liu2023evaluating, sadat2023delucionqa, niu2024ragtruth, ji2024anah} and summarization \citep{bao2025faithbench}.
Despite these advances, existing benchmarks face a fundamental challenge: annotation ambiguity due to the ill-defined boundary of permissible common sense in model outputs, as noted by \citet{seo2025verifying}.
The ambiguity leads to inconsistent evaluations, undermining the reliability of faithfulness assessments.

\begin{table}
    \centering
    \begin{tabularx}{\linewidth}{X}
    \toprule
    \sethlcolor{lime!70}
    \small\textbf{\textsc{Document:}} at the grand old age of 75 , jack nicklaus is still capable of hitting aces \ldots \hl{nicklaus became the youngest person to }\textcolor{red}{\hl{wear a green jacket in 1963 , and collected his sixth}}\hl{ in 1986 .} he is one of five men to complete the career grand slam , an accolade which favourite rory mcilroy can achieve if he wins his third major in succession .\\
    \sethlcolor{cyan!30}
    \small\textbf{\textsc{Response:}} \ldots Villegas made one on the fourth hole like Nicklaus and another on the eighth, but he lost out to Kevin Streelman in a play-off. \hl{Jack Nicklaus is a renowned golfer, having }\textcolor{red}{\hl{won the Masters Tournament six times}}\hl{, including being the youngest person to wear a green jacket in 1963.} \ldots \\
    \hline
    %\textsc{Sentence:} Jack Nicklaus is a renowned golfer, having won the Masters Tournament six times, including being the youngest person to wear a green jacket in 1963. \\
    \small\textbf{\textsc{Label:}} Out-Dependent \\
    %\textsc{Evidence:} nicklaus became the youngest person to wear a green jacket in 1963 , and collected his sixth in 1986 . \\
    \small\textbf{\textsc{Reason:}} It requires external world knowledge (see {https://en.wikipedia.org/wiki/Augusta\_National\_Golf\_Club}) to interpret wearing a green jacket as winning a Masters Tournament championship. \\
    \bottomrule
    \end{tabularx}
    \caption{An example from the \texttt{Out-Dependent} category, where the target sentence and the evidence are highlighted in \sethlcolor{cyan!30}\hl{blue} and \sethlcolor{lime!70} \hl{green}, respectively. The key segment that drives the annotation decision is in \textcolor{red}{red}.}
    \label{tab:not-sure-example}
\end{table}

Specifically, current unfaithfulness detection benchmarks do not consider the annotators' divergent understandings of the model outputs.
%typically assume a uniform standard for judging whether a document supports a claim.
For instance, the widely adopted AIS framework \citep{rashkin2023measuring} relies on an idealized ``generic hearer,'' yet fails to account for annotators' varying cultural backgrounds and domain expertise.
As also reported by \citet{seo2025verifying}, we observe that such ambiguity manifests in existing benchmarks -- for instance, a golf-related claim that interprets ``the green jacket'' as ``the Masters Tournament winner'' (see Table~\ref{tab:not-sure-example}) can be considered fully supported only by those familiar with golf.
While FaithBench \citep{bao2025faithbench} attempts to address this by introducing categories \texttt{Benign} and \texttt{Questionable} for ambiguous cases, their definitions remain vague and do not ensure annotation quality (see Sec.~\ref{sec:related-work} for more details).

To overcome these limitations, we propose a novel annotation framework for LLM faithfulness that clearly labels the ambiguity.
Our key innovation is the rigorous definitions of two intermediate categories, \texttt{Out-Dependent} and \texttt{Ambiguous}, to capture cases where verification depends on external knowledge and on different interpretations.
These definitions offer an efficient alternative to extensive crowdsourcing efforts, such as those employed by \citet{glockner2024ambifc}.
Using this framework, we construct VeriGray, a novel unfaithfulness detection benchmark for summarization, enabling more objective and granular evaluations.
Analysis of our benchmark reveals that even SOTA LLMs are prone to hallucination on the summarization task, and a substantial proportion of generated sentences fall into the \texttt{Out-Dependent} category, underscoring the importance of resolving annotation ambiguity in unfaithfulness detection benchmarks.
Moreover, evaluations on VeriGray show that current detection methods face significant challenges, especially in identifying \texttt{Out-Dependent} and \texttt{Ambiguous} cases, pointing to substantial room for future improvement.

Our contributions are threefold:

\begin{enumerate}
\item \textbf{Framework}: We design a faithfulness annotation framework that clearly labels annotation ambiguity, using the rigorous definitions of class \texttt{Out-Dependent} and \texttt{Ambiguous}.

\item \textbf{Benchmark}: Following the framework, we build an unfaithfulness detection benchmark, VeriGray, annotating 2,044 sentences. An analysis of this benchmark shows a substantial proportion of \texttt{Out-Dependent} sentences, underscoring the importance of addressing annotation ambiguity.

\item \textbf{Analysis}: We show that SOTA LLMs are still prone to unfaithfulness, and that current detection methods face significant challenges, underscoring the need for continued progress in both detecting and mitigating unfaithfulness.
\end{enumerate}

\begin{table*}[thb]
\centering
\resizebox{\textwidth}{!}{
\begin{tabular}{@{}lccccc@{}}
\toprule 
\multirow{2}{*}{\textbf{Benchmark}} & \multirow{2}{*}{\textbf{Generator}} & \textbf{Annotation} & \multicolumn{2}{c}{\textbf{Ambiguity Considered}} & \multirow{2}{*}{\textbf{Annotator}} \\
\cline{4-5}
& & \textbf{Granularity} & \texttt{Knowledge} & \texttt{Linguistic} & \\ \midrule
\citet{cao2021cliff} & BART, PEGASUS &  span-level & \textcolor{green}{\ding{52}} & \textcolor{red}{\ding{56}} & experts \\ 
BEGIN \citep{dziri2022evaluating} & T5-base, GPT2-base & response-level & \textcolor{red}{\ding{56}} & \textcolor{red}{\ding{56}} & human \\ 
VERI-GRAN \citep{liu2023evaluating} & BingChat, NeevaAI, etc. & citation-level & \textcolor{red}{\ding{56}} & \textcolor{red}{\ding{56}} & crowd source \\
HaluEval \citep{li2023halueval} & ChatGPT & response-level & \textcolor{red}{\ding{56}} & \textcolor{red}{\ding{56}} & ChatGPT \\
DelucionQA \citep{sadat2023delucionqa} & ChatGPT & sentence-level & \textcolor{red}{\ding{56}} & \textcolor{red}{\ding{56}} & crowd source \\
ExpertQA \cite{malaviya2024expertqa} & GPT-4, BingChat & sentence-level   & \textcolor{red}{\ding{56}} & \textcolor{red}{\ding{56}} & crowd source \& experts \\
ANAH \citep{ji2024anah} & GPT-3.5 & sentence-level & \textcolor{red}{\ding{56}} & \textcolor{red}{\ding{56}} & human assisted by GPT-4 \\
RAGTruth \citep{niu2024ragtruth} & GPT-4, GPT-3.5, etc. & word-level & \textcolor{red}{\ding{56}} & \textcolor{red}{\ding{56}} & crowd source \\
AmbiFC \citep{glockner2024ambifc} & Human & sentence-level & \textcolor{red}{\ding{56}} & \textcolor{green}{\ding{52}} & crowd source \\
FaithBench \citep{bao2025faithbench} & GPT-4o, GPT-3.5, etc. & span-level & \textcolor{green}{\ding{52}}$^*$ & \textcolor{green}{\ding{52}}$^*$ & graduate students \\
VeriGray (Ours) & GPT-5, DeepSeek-V3, etc. & sentence-level & \textcolor{green}{\ding{52}} & \textcolor{green}{\ding{52}} & graduate students \\
\bottomrule
\end{tabular}
}
\caption{Recent benchmarks for unfaithfulness detection. \textcolor{green}{\ding{52}}$^*$: ambiguity is considered but vaguely defined. \texttt{Knowledge} represents ambiguity induced by external knowledge. \texttt{Linguistic} represents linguistic ambiguity.}
\label{tab:benchmarks}
\end{table*}

\section{Related Work}
\label{sec:related-work}

\noindent\textbf{Unfaithfulness Detection Benchmarks.}
A key limitation in existing benchmarks for unfaithfulness detection is annotation ambiguity, as noted by \citet{seo2025verifying} (see Table~\ref{tab:benchmarks} for a summary).
Most benchmarks overlook the potential differences in cultural backgrounds and domain expertise among annotators.
For instance, the widely used AIS annotation framework \citep{rashkin2023measuring} assumes a ``generic hearer'' as the annotator.
However, in practice, annotators can be diverse, and the annotation guidelines often fail to specify the extent to which common sense or external knowledge can be used to verify LLMs' generation, leading to ambiguity in annotation.
To illustrate, consider the example presented in Table~\ref{tab:not-sure-example}.

Several benchmarks have attempted to address the ambiguity of unfaithfulness annotation, including the one introduced by \citet{cao2021cliff}, AmbiFC \citep{glockner2024ambifc}, and FaithBench \citep{bao2025faithbench}, but each has notable limitations. 
As far as we know, \citet{cao2021cliff} were the first to consider ambiguity stemming from external knowledge and constructed a corresponding benchmark.
Unfortunately, their work relied on old summarization models such as BART \citep{lewis2020bart} and PEGASUS \citep{zhang2020pegasus}, and their insights have not been widely adopted in subsequent research.
AmbiFC, on the other hand, focuses solely on linguistic ambiguity, not covering ambiguity induced by external knowledge.
FaithBench attempts to handle ambiguity by introducing two label categories -- \texttt{Benign} and \texttt{Questionable} -- which refer, respectively, to cases that are ``hallucination, but supported by world knowledge, common sense, or logical reasoning, such that \textit{a reader finds it acceptable or welcomed}'' and cases where ``classification may differ \textit{depending on whom you ask}'', respectively.
Yet these definitions remain subjective and difficult to operationalize consistently.
As shown in Table~\ref{tab:faithbench-bad-cases} of Appendix~\ref{sec:faithbench-annotation-error}, several annotation errors are associated with these two categories.
In contrast to FaithBench, we propose a more rigorous annotation framework designed to systematically address this issue.

\noindent\textbf{Automatic Ambiguity Detection.} \citet{seo2025verifying} identify annotation errors and ambiguity in existing fact-checking benchmarks and propose an automatic detection method that leverages the discrepancies between the human annotation and predictions from multiple LLM-as-a-judge assessments.
The ambiguity patterns they identify fall into four categories, which can be summarized by two categories: ambiguity related to external-knowledge-related and linguistics-related ambiguity, which align with our \texttt{Out-Dependent} and \texttt{Ambiguous} classes, respectively.
However, the automatic detector cannot distinguish between annotation ambiguity and annotation errors, which can result in low precision for ambiguity detection.
In this paper, we employ this ambiguity detector as part of our annotation refinement process.

\section{Annotation Framework}

\begin{table*}[thb]
    \centering
    {\small
    \sethlcolor{lime!70}
    \begin{tabularx}{\textwidth}{>{\hsize=1.2\hsize}X>{\hsize=0.8\hsize}X>{\hsize=1\hsize}X}
    \toprule
    \textbf{Document} & \textbf{Target Sentence} & \textbf{Annotation} \\
    \midrule
    %\sethlcolor{lime!70}\hl{}
    The ``black box'' of the Su-24 jet was officially opened in Moscow on Friday in front of journalists and diplomats. \ldots The Su-24 was shot down by F-16 fighters on 24 November. \ldots
    &  %\sethlcolor{cyan!30}\hl{}
    The ``black box'' of the Su-24 jet, shot down on November 24th, was opened in Moscow on Friday. 
    & \texttt{Explicitly-Supported}  \\
    \midrule
    the city of eugene in the united states has been awarded the rights to host the 2021 iaaf world championships . \ldots the 2007 world championships held in osaka , japan was also awarded without a bidding process . \ldots
    & \textcolor{red}{This is not the first time a World Championships was awarded without a bidding process,} as the 2007 event in Osaka, Japan was also granted in this manner.
    & \texttt{Implicitly-Supported.} \textsc{Reason:} The document supports the claim but does not explicitly mention the highlighted text. \\
    \midrule
    \ldots the ace was nicklaus ' first on the hallowed turf of augusta , but the veteran saw his achievement eclipsed by that of camilo villegas on \textcolor{red}{a day which saw five hole-in-ones , levelling the record from 2002 .} \ldots
    & Despite his achievement, Camilo Villegas stole the show by \textcolor{red}{scoring two holes-in-one, tying the record.}
    & \texttt{Fabricated.} \textsc{Reason:} The record in the document does not refer to the two ace of Villegas.\\
    \midrule
    conor mcgregor is gearing up for a featherweight title challenge against jose aldo on july 11 \ldots \textcolor{red}{aldo makes the eighth defence of his belt} against the irish fighter in las vegas , but mcgregor claimed last week that the man he is challenging lacks the same motivation as him .
    & \textcolor{red}{This is McGregor's eighth defense of the title}, and the two fighters have a contentious relationship.
    & \texttt{Contradicting.} \textsc{Reason:} It's Aldso's 8th defense.\\
    \midrule
    the zhangzhou plant was slated for xiamen \ldots however , \textcolor{red}{it provoked an angry backlash in 2007} due to pollution concerns and \textcolor{red}{prompted the local government to relocate the factory} to its current, more remote location.
    & This plant had experienced a previous explosion in 2013 and \textcolor{red}{was relocated} from Xiamen due to pollution concerns and \textcolor{red}{protests in 2007}. 
    & \texttt{Ambiguous (Explicitly-Supported / Fabricated).} \textsc{Reason:} While the sentence is accurate when ``in 2007'' modifies ``protests,'' the alternative reading -- that it specifies the time of relocation -- lacks support.\\
    \bottomrule
    \end{tabularx}
    }
    \caption{Examples of our taxonomy (see Table~\ref{tab:not-sure-example} for the example of \texttt{Out-Dependent}). The key segment that drives the annotation decision is in \textcolor{red}{red}.}
    \label{tab:examples}
\end{table*}
%\subsection{Annotation Taxonomy}

\begin{figure}[thb]
    \centering
    \includegraphics[width=0.8\linewidth]{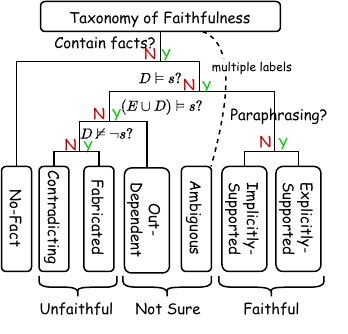}
    \caption{The decision tree for annotating unfaithfulness. The \texttt{Ambiguous} label is not assigned by a single run of the decision tree. Instead, it is applied when the procedure yields multiple labels for different interpretations of the same sentence.}
    \label{fig:taxonomy}
\end{figure}

Let $D$ be the source document and $E$ be external world knowledge (excluding lexical, syntactic, semantic, and pragmatic knowledge of languages, which are considered as the background knowledge inherent to each document).
Assume that $D$ and $E$ do not contradict.
Given a summary $S$ of $D$, our task is to annotate the faithfulness of each sentence $s$ in $S$. 
We propose a fine-grained labeling framework of faithfulness defined below.
See Figure~\ref{fig:taxonomy} for an overview and Table~\ref{tab:examples} for examples.

\noindent\textbf{No-Fact Class.} As identifying verification-worthy sentences is a prerequisite of checking the facts in the generated text \citep{liu2023evaluating}, we first categorize non-verification-worthy sentences as the \texttt{No-Fact} class.
This category includes sentences that do not convey any factual content, e.g., ``would you like to learn more?''.

\noindent\textbf{Faithful Classes.} We observe that a large amount of the LLM-generated summary sentences are paraphrases of the original document, making them straightforward to verify.
Motivated by \textit{selective prediction} \citep{chow1957optimum, geifman2017selective}, we argue that this presents an opportunity for the automatic detectors to accurately classify such easy cases, leaving the more challenging minority of samples to be verified by human experts.
This approach can help reduce the cost of manual fact-checking.
Additionally, annotators can assign a special label to these sentences to indicate their high confidence in their faithfulness.
Accordingly, we classify such sentences as \texttt{Explicitly-Supported} and leave the remaining supported sentences to \texttt{Implicitly-Supported}.
We formalize the definitions as follows. 

\begin{definition}[Explicitly-Supported]
    A sentence $s$ is Explicitly-Supported by document $D$ if and only if i) $s$ is logically implied by $D$, or $D\vDash s$; and ii) For each event, relation, or state in $s$, there exists a semantically equivalent event, relation, or state in $D$.
\end{definition}

\begin{definition}[Implicitly-Supported]
    A sentence $s$ is Implicitly-Supported by document $D$ if and only if $D \vDash s$, and $s$ is not Explicitly-Supported by $D$.
\end{definition}

\noindent\textbf{Unfaithful Classes.} According to the severity of unfaithfulness, we divide unfaithfulness into two classes: \texttt{Contradicting} and \texttt{Fabricated}, which correspond respectively to the Contradiction and Neutral classes of in Natural Language Inference (NLI) task \citep{nie2020adversarial}.
\texttt{Contradicting} aligns directly with the standard Contradiction category in NLI.
However, we define \texttt{Fabricated} with a key distinction from Neutral: it excludes sentences that are supported by external knowledge, as such sentences are instead classified as \texttt{Out-Dependent} (discussed later).
The formal definitions are as follows.

\begin{definition}[Contradicting]
    A sentence $s$ is Contradicting with document $D$ if and only if $s$ logically contradicts $D$, i.e., $D\vDash \neg s$.
\end{definition}

\begin{definition}[Fabricated]
    A sentence $s$ is Fabricated with respect to document $D$ if and only if: $s$ does not contradict $D$, and is neither logically implied by the document nor supported by external world knowledge, i.e., $D\nvDash s, \neg s$ and $(E\cup D)\nvDash s$.
\end{definition}

\noindent\textbf{Not Sure Classes.} This category addresses cases where annotators may reach different judgments due to inherent ambiguities.
Empirically, we identify two common patterns: (1) vague boundaries of permissible external knowledge in generated text, and (2) generated text or source documents that support multiple interpretations with different faithfulness classes.
For the first pattern, we define the \texttt{Out-Dependent} class as follows.
\begin{definition}[Out-Dependent]
    A sentence $s$ is Out-Dependent with respect to document $D$ if and only if $s$ is not logically implied by the document alone, but is entailed by the document combined with external world knowledge, i.e., $D\nvDash s$, and $(E\cup D)\vDash s$.
\end{definition}

At present, the classification for sentences with a single interpretation is complete (as illustrated in Figure~\ref{fig:taxonomy}).
We now turn to the cases involving multiple interpretations arising from linguistic ambiguity, i.e., the second pattern noted above.
For such cases, we preserve all plausible classes associated with each interpretation and collectively designate them as the \texttt{Ambiguous} class.

%\begin{table}[thb]
%    \centering
%    \resizebox{\linewidth}{!}{
%    \begin{tabular}{lcccc}
%    \toprule
%         & \begin{tabular}{c}
%            \textbf{Verbatim or} \\
%            \textbf{Paraphrasing}
%         \end{tabular} & $D\vDash s$ & $D\vDash \neg s$ & $(E\cup D)\vDash s$ \\
%    \hline
%    \texttt{EXS} & \ding{52} & \ding{52} & \ding{56} & \ding{52} \\
%    \texttt{IMS} & \ding{56} & \ding{52} & \ding{56} & \ding{52} \\
%    \texttt{OUT}       & \ding{56} & \ding{56} & \ding{56} & \ding{52} \\
%    \texttt{FAB}   & \ding{56} & \ding{56} & \ding{56} & \ding{56} \\
%    \texttt{INC} & \ding{56} & \ding{56} & \ding{52} & \ding{56} \\
%    \bottomrule
%    \end{tabular}
%    }
%    \caption{Summarization of the taxonomy, where \texttt{EXS/IMS/OUT/FAB/INC} represent \texttt{Explicitly-Supported / Implicitly-Supported / Out-Dependent / Fabricated / Contradicting}, respectively.}
%    \label{tab:taxonomy}
%\end{table}

\section{VeriGray: A dataset of Objective Unfaithfulness Annotation}

Considering that summarization is a typical scenario for unfaithfulness detection \citep{scialom2021questeval, pagnoni2021understanding, niu2024ragtruth, bao2025faithbench}, we build a summarization dataset annotated with unfaithfulness labels, as detailed below.

\subsection{Data Collection}

We collected the documents from FaithBench \citep{bao2025faithbench}, whose passages for summarization come from various NLI, fact-checking, and summarization datasets.
To exclude NLI instances, we removed instances where the document is shorter than twice the length of its corresponding summary. 
We also removed self-contradictory documents.
The remaining documents, along with their LLM-generated summaries, were retained, but original faithfulness annotations were removed.
Subsequently, we employed more recent LLMs, including GPT-5 \cite{openai2025introducing}, DeepSeek V3-0324 \citep{liu2024deepseek}, and Qwen3-8B \citep{yang2025qwen3}, to generate additional summaries for each document.
Following the setup in FaithBench, we used the summarizer prompt of Vectara's Hallucination Leaderboard \cite{hughes2023vectara}.
The full prompt can be found in Appendix~\ref{sec:summary-generating-details}.
We set the temperatures of DeepSeek-V3 and Qwen3-8B to 0.3 and 0.6, respectively (the temperature of GPT-5 was not adjustable via the API).
After appending the newly generated summaries, our dataset comprised 412 summaries, containing a total of 2044 sentences.
Each summary was segmented into sentences using NLTK \citep{bird2006nltk}, with manual corrections applied by annotators as needed.

\subsection{Human Annotation}
\label{sec:human-annotation}

Each sentence in the generated summaries was evaluated for faithfulness by two human annotators, following the annotation framework outlined in Figure \ref{fig:taxonomy}.
Specifically, if a sentence was not directly supported by the source document, annotators further assessed whether it could be verified using a combination of the document and external knowledge. External knowledge was retrieved via Bing Search.
In cases where a sentence was labeled as \texttt{Out-Dependent}, annotators were required to cite URLs from reliable external sources—such as Wikipedia or mainstream media websites. Conversely, if a sentence was labeled as Unfaithful, annotators had to provide evidence from the source document along with a clear reasoning process.
To ensure reproducibility, all relevant web pages referenced during the annotation process were saved and included as attachments\footnote{They are also open sourced at \url{https://huggingface.co/datasets/Ding-Qiang/veri-gray}.} to the dataset.
For more details of the annotation guidelines, please refer to Appendix \ref{sec:annotation-special-cases}.

\noindent\textbf{Annotators.} The annotation team consisted of two graduate students whose expertise lies in natural language processing, both of whom have previously published on trustworthy AI at top-tier ML/NLP conferences. 
All annotators were aware that the annotated data would be made publicly available.

\noindent\textbf{Attribution-assisted Annotation.} We observed that the most time-consuming aspect of faithfulness annotation is locating relevant evidence spans in the document.
To streamline this process, we integrated an attention-based fine-grained attribution method \citep{ding2025attention} into a web-based annotation tool.
When an annotator selects a target sentence, the attribution module automatically highlights relevant text segments in the document, facilitating efficient evidence identification.

\noindent\textbf{Quality Assurance.} %Each instance was initially labeled by one annotator and subsequently reviewed by another, in conjunction with a modified version of the automatic annotation error detector from \citet{seo2025verifying}.
The quality assurance process was conducted in multiple stages to ensure high quality. 
Firstly, before annotation, all annotators completed a training phase that included reviewing canonical examples and passing a quiz on challenging cases.
Secondly, during annotation, annotators selected not only the label but also intermediate decision options such as \texttt{Contains Fast}, \texttt{Is Ambiguous}, and \texttt{Is Supported by Doc}.
The system automatically verified the consistency between these intermediate choices and the final label, and only consistent annotations could be submitted.
Thirdly, after annotation, each instance underwent successive reviews by a second annotator.
Fourthly, instances not classified as \texttt{Ambiguous} or \texttt{Out-Dependent} were processed by a modified version\footnote{The original error detector runs four LLM-as-a-judge models, with its final prediction confirmed by a second round of LLM-as-a-judge. Our modification introduced a new LLM-as-a-judge model of GPT-5 and omitted the second round as it empirically reduced error detection recall compared to the first round alone.} of the automatic annotation error detector proposed by \citet{seo2025verifying}.
Specifically, the modified detector runs five LLM-as-a-judge models (using o3-mini, GPT-4o, Gemini 2.0-Flash, Llama3.1 405B, and GPT-5), each providing their individual faithfulness predictions (\texttt{Attributable}, \texttt{Not Attributable}, or \texttt{Contradictory}). Potential annotation errors were flagged whenever any LLM's prediction disagreed with the annotated label.
Finally, error detection results were reviewed by the annotators to produce the final annotations.

\begin{figure*}[thb]
    \centering
    \includegraphics[width=\textwidth]{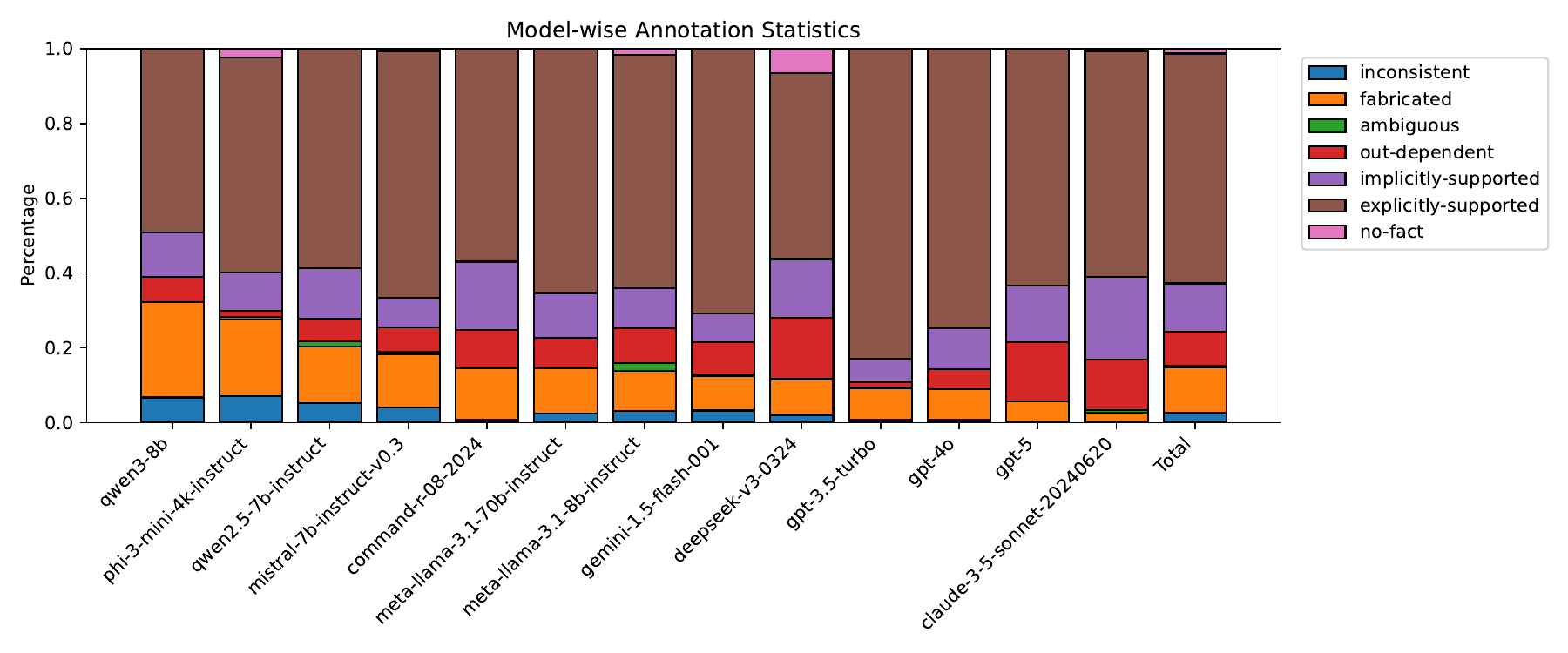}
    \caption{Model-wise annotation statistics, where the models are arranged from left to right according to the descending order of the proportion of unfaithful classes.}
    \label{fig:annotation-statistics}
\end{figure*}

\subsection{Dataset Analysis}

Here, we investigate the distribution of unfaithfulness classes in our benchmark\footnote{We also explore the sequential dependency of unfaithfulness in Appendix \ref{sec:sequential-dependency} to see if the phenomenon of \textit{hallucination snwoballing} \citep{zhang2024language} exists in our benchmark.}.
%\noindent\textbf{Distribution of Unfaithfulness Classes.}
Figure~\ref{fig:annotation-statistics} shows the sentence-level category breakdown across different models.
Overall, the majority are Faithful (74\%), with \texttt{Explicitly-Supported} sentences (61\%) substantially outnumbering \texttt{Implicitly-Supported} sentences (13\%).
Unfaithful sentences constitute approximately 15\% of the total, most of which are \texttt{Fabricated} (12\%).
A notable proportion of sentences fall under the Not Sure category, primarily as \texttt{Out-Dependent} (9\%), while fewer than 1\% are labeled as \texttt{Ambiguous}.
Across models, DeepSeek-V3-0324 and GPT-5 yield the highest proportion of Not Sure sentences (16\%).
Meanwhile, Claude-3.5-Sonnet generates the fewest Unfaithful sentences (3\%), surpassing the more recent GPT-5 (6\%), whereas Qwen3-8B generates the most (32\%), reflecting the comparative success of proprietary models in maintaining faithfulness.

%\begin{table}[thb]
%\centering
%\resizebox{\linewidth}{!}{
%\begin{tabular}{lcc}

%\toprule
% & \textbf{To Faithful} & \textbf{To Unfaithful} \\
%\midrule
%    Explicitly-Supported & $0.80\pm0.02$ & $0.13\pm0.02$ \\
%    Implicitly-Supported & $0.75\pm0.06$ & $0.15\pm0.05$ \\
%    Fabricated & $0.71\pm0.07$ & $0.21\pm0.06$ \\
%    Inconsistent & $0.74\pm0.13$ & $0.21\pm0.12$ \\
%\bottomrule
%\end{tabular}
%}
%\caption{Estimated transition probabilities $\hat p$ (with confidence of 95\%) of faithfulness polarity. Rows and columns correspond to source and target classes, respectively. Target fine-grained labels are merged into Faithful/Unfaithful/Others classes to analyze sequential dependency at the polarity level.}
%\label{fig:transition-probabilities}
%\end{table}

\section{Experiments}

We now employ VeriGray to assess a range of unfaithfulness detection baselines.
Section~\ref{sec:baselines} outlines the baselines to be evaluated.
Section~\ref{sec:evaluation} details the evaluation protocols adapted for prior unfaithfulness detectors, which operate in output spaces different from our taxonomy.
The results are then presented in Section~\ref{sec:results}.

\subsection{Baselines}
\label{sec:baselines}

\noindent\textbf{Fine-tuned LLMs.} The fine-tuned LLMs comprise \text{Minicheck 7B}\footnote{\url{https://huggingface.co/bespokelabs/Bespoke-MiniCheck-7B}} \citep{tang2024minicheck}, \text{Lynx 8B Instruct v1.1}\footnote{\url{https://huggingface.co/PatronusAI/Llama-3-Patronus-Lynx-8B-Instruct-v1.1}}, \text{Lynx 70B Instruct}\footnote{\url{https://huggingface.co/PatronusAI/Llama-3-Patronus-Lynx-70B-Instruct}} \citep{ravi2024lynx}, and \text{ANAH-v2}\footnote{\url{https://huggingface.co/opencompass/anah-v2}} \citep{gu2024anahv2}, where Minicheck and Lynx models are binary classifiers, and ANAH-v2 is a three-way classifier.

\noindent\textbf{Zero-shot LLMs.} Zero-shot LLMs are prompted with the definitions from our taxonomy (see Appendix \ref{sec:zero-shot-details} for the prompt).
The evaluated LLMs include \text{GPT-5}, \text{DeepSeek-R1} \citep{guo2025deepseekr1}, \text{DeepSeek-V3}, \text{Qwen3 235B A22B} \citep{yang2025qwen3}, and \text{QwQ-32B} \citep{team2025qwq}.
We also included GPT-5 enhanced with RAG \citep{lewis2020retrieval} -- hereafter referred to as GPT-5 + RAG -- to provide the LLM with access to external world knowledge and help verify \texttt{Out-Dependent} examples.
The retrieval corpus was the webpages collected during annotation (see Section \ref{sec:human-annotation}). 
For more details, please refer to Appendix \ref{sec:zero-shot-details}.

\noindent\textbf{Others.} The other baselines include two model-internals-based methods, \text{LLM-Check} \citep{sriramanan2024llmcheck} and ReDeEP \citep{sun2024redeep}, a probability-based method \text{CCP} (Claim Conditioned Probability) \citep{fadeeva2024factchecing}, and a retrieve-then-verify approach \text{InFusE} \citep{zhang2024finegrained}.
Implementation details for all baselines are provided in Appendix \ref{sec:baselines-implementation}.

\subsection{Evaluation Protocol}
\label{sec:evaluation}

%\noindent\textbf{Evaluating Previous Work.} 
Since the label space in our benchmark differs from previous work, where hallucination detection is typically treated as a binary or three-way classification, we propose two evaluation protocols to overcome this difficulty. 
Following \citet{seo2025verifying}, the first protocol removes Not Sure and \texttt{No-Fact} instances and evaluates detectors on the remaining data, with all classes merged into two categories: Faithful and Unfaithful.
This merging involves two aspects: merging annotations and merging predictions.
For annotations, \texttt{Explicitly-Supported} and \texttt{Implicitly-Supported} are merged into the Faithful class, while \texttt{Fabricated} and \texttt{Contradicting} form the Unfaithful class.
Predictions are then aligned accordingly: for NLI-style detectors, Entailment is mapped to Faithful, and Neutral and Contradictory to Unfaithful; for zero-shot LLMs, classes with a faithfulness degree not less than a threshold (defined later) are considered Faithful, with the rest as Unfaithful.
Here, the \textit{order of faithfulness degree} is defined as: $\texttt{Contradicting} \prec \texttt{Fabricated} \prec \texttt{Ambiguous} \prec \texttt{No-Fact} \prec \texttt{Out-Dependent} \prec \texttt{Implicitly-Supported} \prec \texttt{Explicitly-Supported}$.
The threshold is currently set to the \texttt{Implicitly-Supported} class, with alternative thresholds explored in the selective prediction part of Section~\ref{sec:results}.
After merging, we report \textbf{balanced accuracy}, along with \textbf{hallucination detection recall, precision, and F1} (i.e., the recall, precision, and F1 for the Unfaithful class).

To fully utilize the fine-grained annotations, we propose another protocol that evaluates the ranking quality of the hallucination detectors.
We consider the full dataset filtering out \text{No-Fact} instances only, denoted as $\mathcal{D}_{\text{fact}} = \{(x_i, y_i) | y_i \neq \texttt{No-Fact}\}$, where $y_i$ is the fine-grained label of instance $x_i$.
For predictions, instead, the output space is merged into Faithful/Unfaithful as before, assigned with a faithfulness degree order of Unfaithful $\prec$ Faithful.
%\begin{equation}
%    \text{F}(y) := \begin{cases}
%        6, \text{ if } y = \text{Explicitly-Supported} \\
%        5, \text{ if } y = \text{Implicitly-Supported} \\
%        4, \text{ if } y = \text{Out-Dependent} \\
%        3, \text{ if } y = \text{No-Fact} \\
%        2, \text{ if } y = \text{Ambiguous} \\
%        1, \text{ if } y = \text{Fabricated} \\
%        0, \text{ if } y = \text{Contradicting}
%    \end{cases}
%    \label{eq:faithfulness-degree-definition}
%\end{equation}
Let $\mathcal{P} := \{(x_i, x_j) | y_i \succ y_j, (x_i, y_i), (x_j, y_j) \in \mathcal{D}_{\text{fact}}\}$.
Inspired by the ranking expression of AUC (Area Under the Curve; Eq. (2.21) in \citet{zhou2021machine}), we propose a novel metric \textbf{ranking loss}:
\begin{align}
L_{\text{rank}} := & \frac{1}{|\mathcal{P}|}\sum_{(x_i, x_j) \in \mathcal{P}} \bigg(\mathbb{I}[f(x_i) \prec f(x_j) ] \notag \\
& + \frac 12 \mathbb{I}[f(x_i) = f(x_j)]\bigg), \label{eq:rank-loss}
\end{align}
where $f$ is the prediction of the hallucination detector. This metric is non-negative and equals zero if and only if $f$ perfectly preserves the faithfulness ranking of $y$. 
%The weighting term $\text{F}(y_i) - \text{F}(y_j)$ is added to penalize faithfulness disorders over distant faithfulness classes.

In addition, for zero-shot LLMs, we evaluate the \textbf{class-wise precision} and \textbf{recall} to investigate the fine-grained unfaithfulness classification performances.

\begin{figure*}[htbp]
    \centering
    \begin{minipage}{0.35\textwidth}
        \centering
        \includegraphics[width=\linewidth]{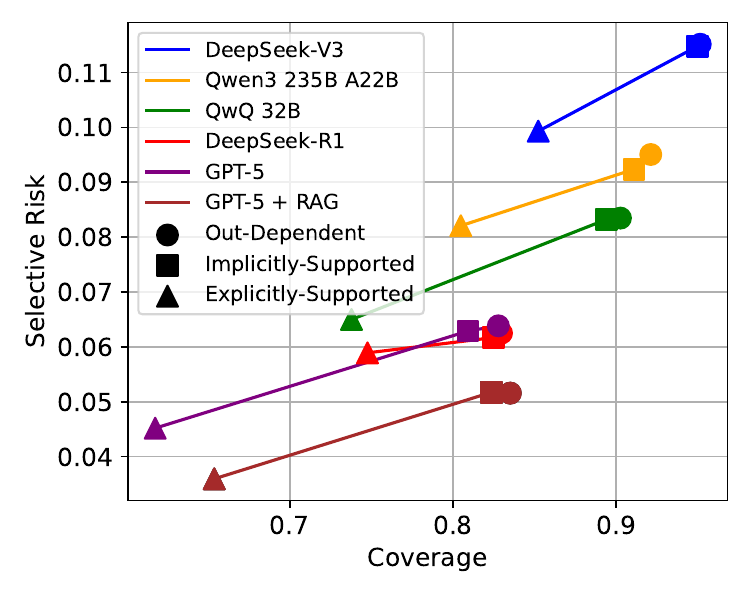}
        \caption{The selective prediction results of zero-shot LLMs on VeriGray. Colors denote different models, and marker shapes denote different confidence thresholds.}
        \label{fig:selective-prediction-results}
    \end{minipage}
    \hfill
    \begin{minipage}{0.63\textwidth}
        \centering
        \includegraphics[width=\linewidth]{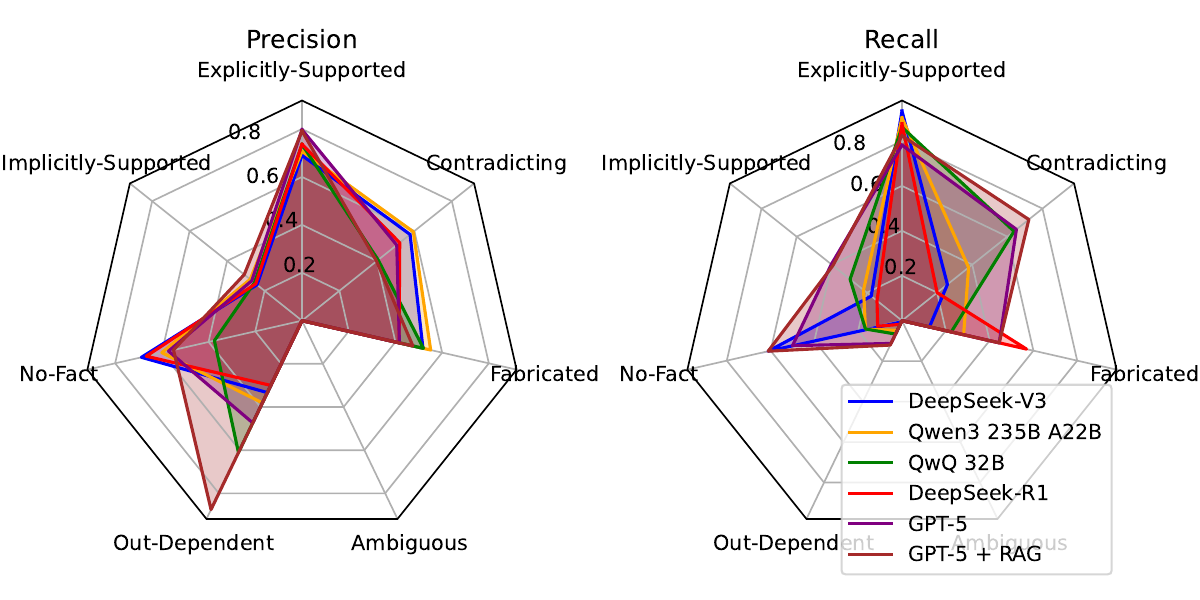}
        \caption{The class-wise precision and recall of zero-shot LLMs. The low recall for \texttt{Ambiguous} may stem from the scarcity of \texttt{Ambiguous} examples.}
        \label{fig:class-wise-pr}
    \end{minipage}
\end{figure*}

\subsection{Results}
\label{sec:results}

\begin{table*}[thb]
\small
\centering
%\resizebox{\textwidth}{!}{
\begin{tabular}{lccccc}
\toprule
& \textbf{BAcc} ($\uparrow$) & \textbf{Hallu. Rec.} ($\uparrow$) & \textbf{Hallu. Prec.} ($\uparrow$) & \textbf{Hallu. F1} ($\uparrow$) & $L_{rank}$ ($\downarrow$) \\  % & \textbf{Macro-F1} ($\uparrow$) \\
\midrule
\rowcolor{gray!20}\multicolumn{6}{c}{\textit{Zero-shot LLMs}} \\
DeepSeek-V3           & 60.4 & 21.3 & \textbf{90.1}   & 34.5 & 44.3 \\
Qwen3 235B A22B       & 68.5 & 39.0 & 80.1            & 52.5 & 40.0 \\
QwQ 32B               & 73.1 & 48.0 & \underline{84.7}& 61.3 & 37.5 \\
DeepSeek-R1           & 79.3 & 64.7 & 68.1 & 66.3     & 33.4 \\
GPT-5                 & \underline{82.5} & \underline{71.0} & 69.8 & \underline{70.4} & \underline{32.1} \\
GPT-5 + RAG           & \textbf{83.6} & \textbf{72.0} & 75.0 & \textbf{73.5} & \textbf{31.9} \\
\rowcolor{gray!20}\multicolumn{6}{c}{\textit{Fine-tuned LLMs}} \\
ANAH-v2               & 53.3 & 11.7 & 31.0 & 16.9 & 46.8 \\  % &    \\
MiniCheck 7B          & 60.8 & 25.7 & 55.4 & 35.1 & 43.2 \\  % &    \\
Lynx 8B Instruct v1.1 & 61.7 & 37.0 & 34.8 & 35.9 & 42.4 \\  % &    \\
Lynx 70B Instruct     & 66.1 & 38.7 & 54.2 & 45.1 & 42.1 \\  % &    \\
%\hline
%\rowcolor{gray!20}\multicolumn{7}{c}{\textit{Model-internals-based Methods}} \\
\rowcolor{gray!20}\multicolumn{6}{c}{\textit{Others}} \\
LLM-Check (Attn Score) & 53.6 & 59.3 & 18.3 & 28.0 & 47.5 \\  % &  \\ 
LLM-Check (Hidden Score)&54.7 & 56.0 & 19.2 & 28.6 & 48.6 \\  % &  \\
ReDeEP                & 57.1 & 66.7 & 20.1 & 30.8 & 43.3 \\
InFusE                & 46.8 & 38.0 & 14.4 & 20.9 & 52.1 \\  % &   \\
CCP                   & 58.9 & 48.0 & 23.8 & 31.9 & 44.2 \\  % &   \\
\bottomrule
\end{tabular}
%}
\caption{The evaluation results (\%) of balanced accuracy (BAcc), hallucination recall (Hallu. Rec.), hallucination precision (Hallu. Prec.), and hallucination F1 (Hallu. F1). The best entries are marked in \textbf{bold}, and the second-best entries are \underline{underlined}.}
\label{tab:results}
\end{table*}

%\begin{table}[tbh]
%    \centering
%    \resizebox{\linewidth}{!}{
%    \begin{tabular}{lcccc}
%        \toprule
%        & \multicolumn{3}{c}{\textbf{Not Sure}} & \multirow{2}{*}{\textbf{Macro-F1}}\\
%        \cmidrule{2-4}
%        & \textbf{Rec.} & \textbf{Prec.} & \textbf{F1} &  \\
%        \midrule
%        GPT-5           & \textbf{9.7}    & \underline{41.5} & \textbf{15.7}   & \textbf{39.3} \\
%        DeepSeek-R1     & 2.3             & 33.3             & 4.3             & 31.2 \\
%        DeepSeek-V3     & 0.6             & 25.0             & 1.1             & 30.1 \\
%        Qwen3 235B A22B & 1.1             & 18.2             & 2.2             & 26.9 \\
%        QwQ 32B         & \underline{5.1} & \textbf{52.9}    & \underline{9.4} & \underline{31.8} \\
%        \bottomrule
%    \end{tabular}
%    }
%    \caption{The evaluation results (\%) of metrics specific to zero-shot LLMs. The best entries are marked in \textbf{bold}, and the second-best entries are \underline{underlined}.}
%    \label{tab:specific-evaluation-results}
%\end{table}

The main evaluation results are shown in Table~\ref{tab:results}.
As shown, the most effective methods are zero-shot LLMs, notably GPT-5 + RAG, which achieve balanced accuracy of 83.6\%, hallucination F1 of 73.5\%, and ranking loss of 31.9\%.
Although zero-shot methods perform well in coarse-grained unfaithfulness detection, their fine-grained detection performances still face significant challenges.

Figure~\ref{fig:class-wise-pr} shows the fine-grained precision and recall of zero-shot LLMs.
Across all models, recalls for \texttt{Ambiguous} and \texttt{Out-Dependent} remain consistently low.  %, suggesting that LLMs have difficulty distinguishing Not Sure instances from others.
Nevertheless, compared to GPT-5, GPT-5 + RAG significantly boosts the precision of \texttt{Out-Dependent} while maintaining the recall, showing the effectiveness of using RAG.
The low recall for \texttt{Ambiguous} may stem from both the scarcity of \texttt{Ambiguous} examples and the models' lack of sensitivity to linguistic ambiguity.
As for \texttt{Out-Dependent}, we found this class was mainly misclassified as faithful classes (see Figure \ref{fig:gpt-5-rag-confusion} in Appendix \ref{sec:confusion}), indicating that the low recall may be due to the LLM detectors' unawareness of using external knowledge from their parametric memory.

\noindent\textbf{Selective Prediction Evaluation.} To investigate whether the \texttt{Explicitly-Supported} prediction serves as a reliable indicator of faithfulness confidence, we frame zero-shot LLM detectors as \textit{confidence estimators} and assess their performance under the framework of \textit{selective prediction} \citep{geifman2017selective}.
In this setting, selective prediction assesses both the \textit{selective risk} -- the hallucination rate for samples whose confidence is greater than or equal to a certain threshold -- and the \textit{coverage}, defined as the proportion of samples with confidence greater than or equal to that threshold.
Here, confidence is defined as the faithfulness degree, whose ordering has been established earlier.
We vary the confidence threshold across \texttt{Out-Dependent}, \texttt{Implicitly-Supported}, and \texttt{Explicitly-Supported} to comprehensively evaluate the selective prediction performance.
For a fixed coverage, a lower selective risk indicates better selective prediction performance.

The selective risk-coverage plots are shown in Figure~\ref{fig:selective-prediction-results}. 
When the threshold transits from \texttt{Out-Dependent} to \texttt{Explicitly-Supported}, selective risk decreases consistently.
GPT-5 + RAG achieves the best tradeoff between selective risk and coverage, reaching below 4\% with coverage over 60\%.
The results highlight the potential of selective prediction in enhancing trustworthiness for unfaithfulness detection.

\section{Conclusion}

This paper presents a novel benchmark, VeriGray, for unfaithfulness detection that systematically addresses long-overlooked issues of knowledge-level and linguistic ambiguity.
Our analysis demonstrates that even state-of-the-art LLMs like GPT-5 exhibit non-trivial rates of unfaithful generation (approximately 6\%), and consistently produce content requiring external knowledge to verify -- validating the need for a dedicated class for such ambiguous cases.
Thus, our benchmark rigorously defines and annotates the ambiguity.
Experiments reveal that zero-shot LLMs are most effective on this benchmark, though far from completely solving the problem.
Specifically, these models, without access to external knowledge, struggle to detect \texttt{Out-Dependent} unfaithfulness.
This indicates a critical path for future research: developing detection methods that are augmented with external knowledge sources.

\section*{Limitations}

Due to the limitation of human labor, we only consider the faithfulness over summarization tasks in our benchmark.
One direction of future work of our paper could be building benchmarks with diverse knowledge-grounded tasks, such as RAG and data-to-text.
Another direction of future work might be to develop methods that can identify when an LLM leverages external world knowledge from its parametric memory, which would enhance the detection of \texttt{Out-Dependent} examples.

\section*{Acknowledgements}

I want to express my sincere gratitude to ZHONG Yang for his careful review and correction of dozens of annotation errors in the dataset, which significantly improved the data quality.

\bibliography{custom}

\appendix

\section{Case Study of Annotation Errors in FaithBench}
\label{sec:faithbench-annotation-error}

Several annotation errors of \texttt{Benign} and \texttt{Questionable} are shown in Table~\ref{tab:faithbench-bad-cases}.
 
\begin{table*}[thb]
    \centering
    {\small
    \sethlcolor{lime!70}
    \begin{tabularx}{\textwidth}{>{\hsize=1.3\hsize}X>{\hsize=1\hsize}X>{\hsize=0.7\hsize}X>{\hsize=1\hsize}X}
    \toprule
    \textbf{Document} & \textbf{Target Sentence} & \textbf{Original Annot.} & \textbf{Our Annotation} \\
    \midrule
    \ldots three British writers, two Americans and an Australian on this year's shortlist. \ldots This year marks is \textcolor{red}{46th year of the Booker Prize}. \ldots
    & Here's a concise summary of the passage: The passage discusses the \textcolor{red}{2014 Man Booker Prize} shortlist and its contenders. 
    & All annotators agreed to \texttt{Consistent} (it seems to be \texttt{Benign} instead).
    & \texttt{Out-Dependent}. \textsc{Reason}: It requires external world knowledge to tell in which year the 46th Booker Prize was awarded. \\
    \midrule
    \ldots  Turkey insists that the jet, \ldots ignored warnings to leave \textcolor{red}{its airspace}. Russia says it was shot down within Syrian airspace and President Vladimir Putin \ldots
    & Turkey claims the jet ignored warnings to leave \textcolor{red}{Syrian airspace}, while Russia asserts it was shot down within Syrian airspace.
    & All annotators agreed to \texttt{Unwanted} (it seems to be \texttt{Questionable} instead).
    & \texttt{Ambiguous (Contradicting / Explicitly-Supported)}. \textsc{Reason}: The annotation depends on how to interpret ``its'' in the document. \\
    \midrule
    \ldots \textcolor{red}{the plant was hit by another explosion in july 2013}, although there were no reports of casualties or toxic leaks at the time . \ldots
    & There was no mention of toxic chemical spills in this incident, but the plant has a history of explosions, with \textcolor{red}{the last one occurring in July 2013}. 
    & Two of three annotators labeled the sentence as \texttt{Questionable}.
    & \texttt{Out-Dependent}. \textsc{Reason}: It requires external knowledge to tell whether the explosion in July 2013 is the most recent one. \\
    \midrule
    \ldots but the veteran saw his achievement eclipsed by that of camilo villegas on \textcolor{red}{a day which saw five hole-in-ones , levelling the record from 2002.} the colombian followed nicklaus with an ace on the fourth and also managed one on the eighth , but he could n't quite land the title \ldots
    & Despite his achievement, Camilo Villegas stole the show by \textcolor{red}{scoring two holes-in-one, tying the record.}
    & One of four annotators labeled this sentence as \texttt{Benign}. Two of four annotators labeled this sentence as \texttt{Questionable}.
    & \texttt{Fabricated}. \textsc{Reason}: The record was not the two aces of Villegas but the five aces on that day. \\
    \bottomrule
    \end{tabularx}
    }
    \caption{Bad cases of FaithBench annotation related to \texttt{Benign} and \texttt{Questionable} classes, where \textbf{Original Annot.} denotes the original annotations. Considering the original annotations are span-level, we aggregate all span-level original annotations whose span is in the target sentence to be compared with our annotations.}
    \label{tab:faithbench-bad-cases}
\end{table*}

\section{Details of Generating the Summaries}
\label{sec:summary-generating-details}

The summarizer prompt is as follows.
Here, the <PASSAGE> is a placeholder for the real document.

\begin{tcolorbox}[title=Summarizer Prompt]
You are a chat bot answering questions using data. You must stick to the answers provided solely by the text in the passage provided. You are asked the question `Provide a concise summary of the following passage, covering the core pieces of information described.' <PASSAGE>'
\end{tcolorbox}

\section{Special Cases of Annotation}
\label{sec:annotation-special-cases}

\noindent\textbf{Quotations and Citations}.
There are many quotations and citations in the collected documents.
Inspired by the conventions of academic paper reading, we consider quotations and citations with clear sources to be reliable.
Therefore, it is appropriate for the summary to state their contents with confidence.
For anonymous citations, we regard them as uncertain facts and consider summaries that do not convey uncertainty as not supported by the documents.

\noindent\textbf{Meta Notes}. For some LLMs, there are many meta notes in the generated text, e.g., ``*(Note: The mention of Mo Farah’s 2016 gold medal is unrelated to the core information and excluded.)*''.
Although these meta notes state some facts, they are not part of the summaries.
Therefore, we label these meta notes as \texttt{No-Fact} to exclude them from evaluation.

\section{Sequential Dependency of Unfaithfulness}
\label{sec:sequential-dependency}

\begin{figure}
    \centering
    \includegraphics[width=\linewidth]{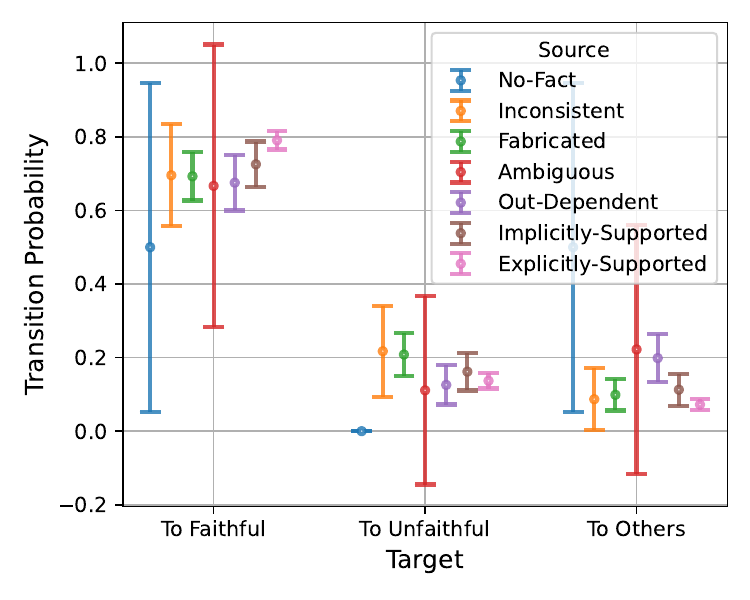}
    \caption{Estimated transition probabilities $\hat p$ (with confidence of 95\%) of faithfulness polarity. Target labels are merged into Faithful/Unfaithful/Others classes to analyze sequential dependency at the polarity level.}
    \label{fig:transition-probabilities}
\end{figure}

\begin{figure*}
    \centering
    \subfigure{
    \includegraphics[width=0.23\textwidth]{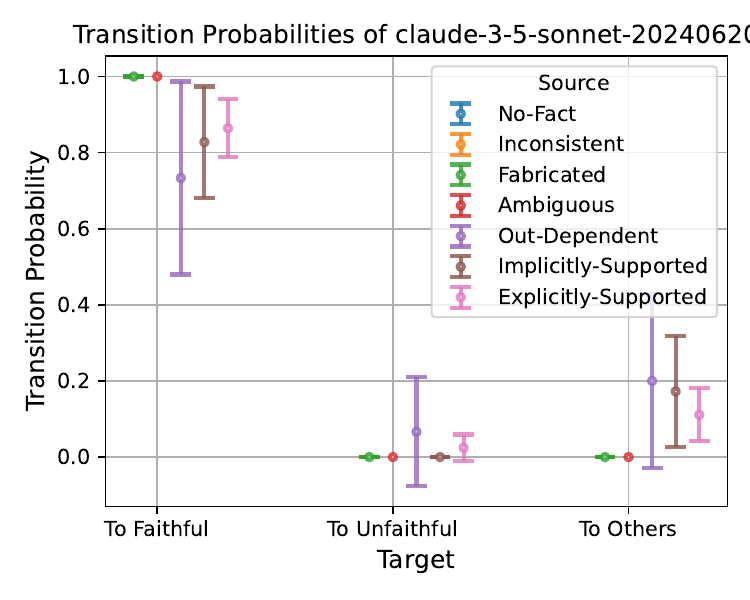}
    }
    \subfigure{
    \includegraphics[width=0.23\textwidth]{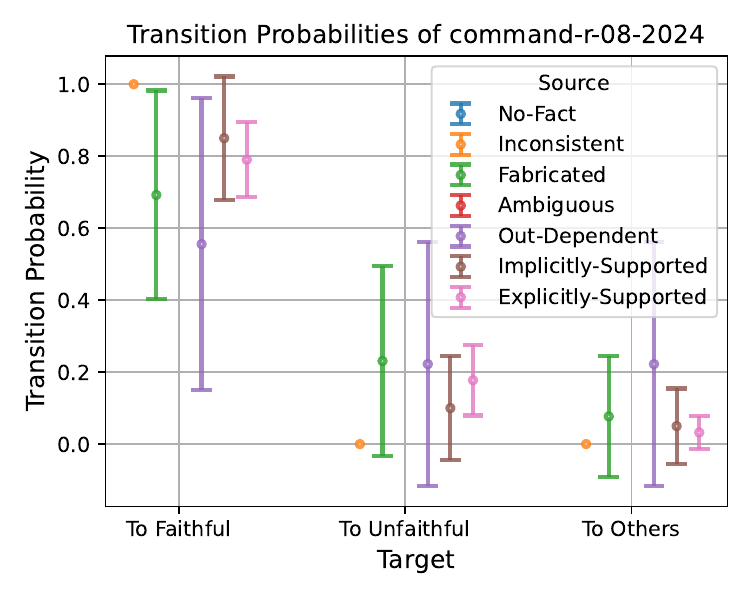}
    }
    \subfigure{
    \includegraphics[width=0.23\textwidth]{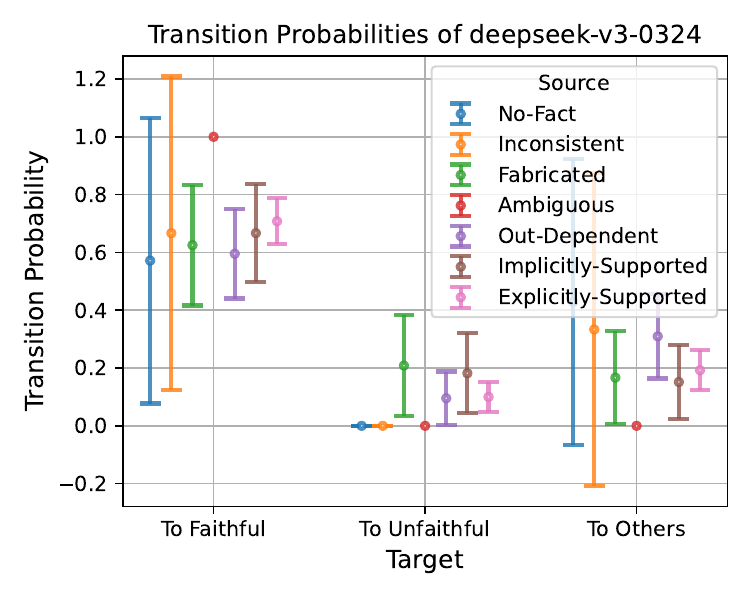}
    }
    \subfigure{
    \includegraphics[width=0.23\textwidth]{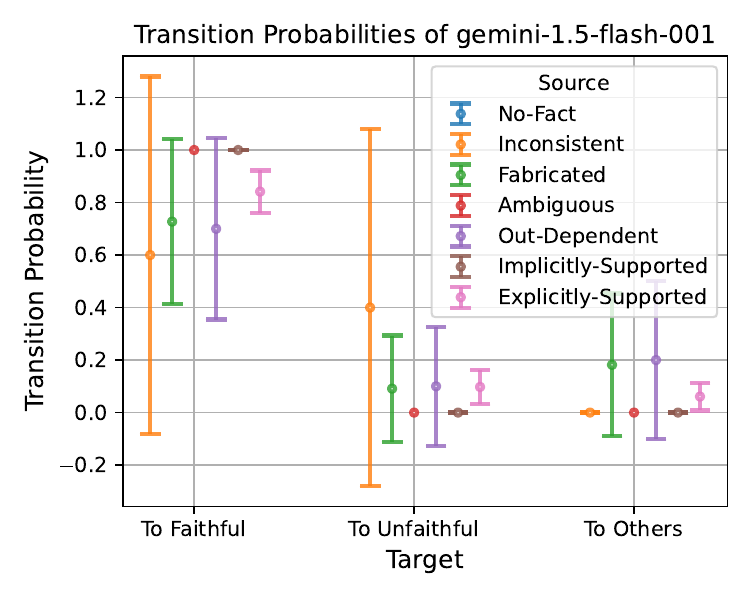}
    }
    \subfigure{
    \includegraphics[width=0.23\textwidth]{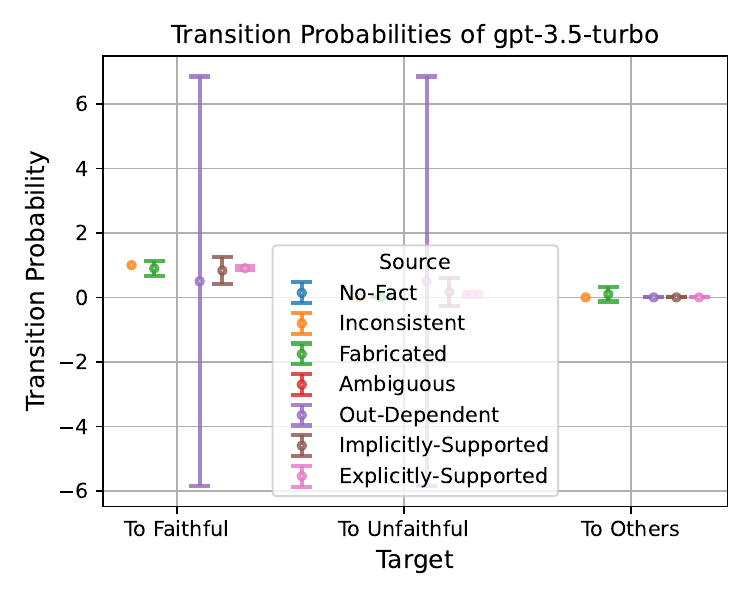}
    }
    \subfigure{
    \includegraphics[width=0.23\textwidth]{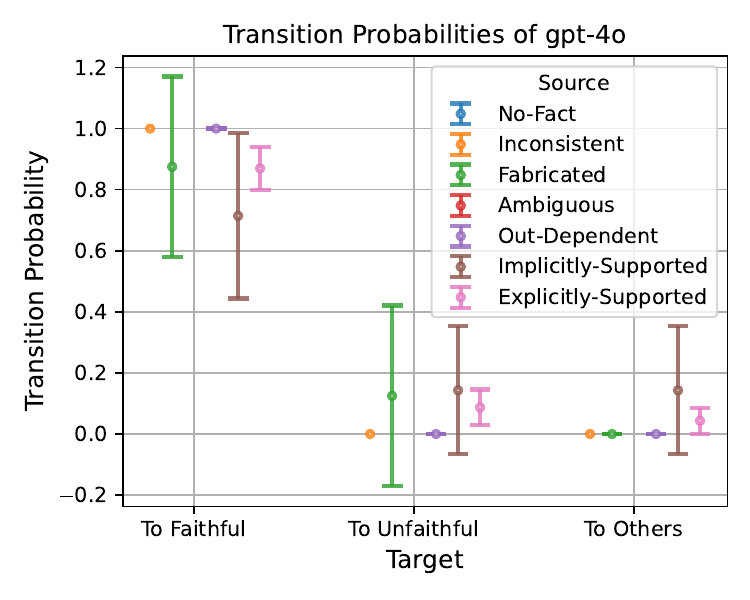}
    }
    \subfigure{
    \includegraphics[width=0.23\textwidth]{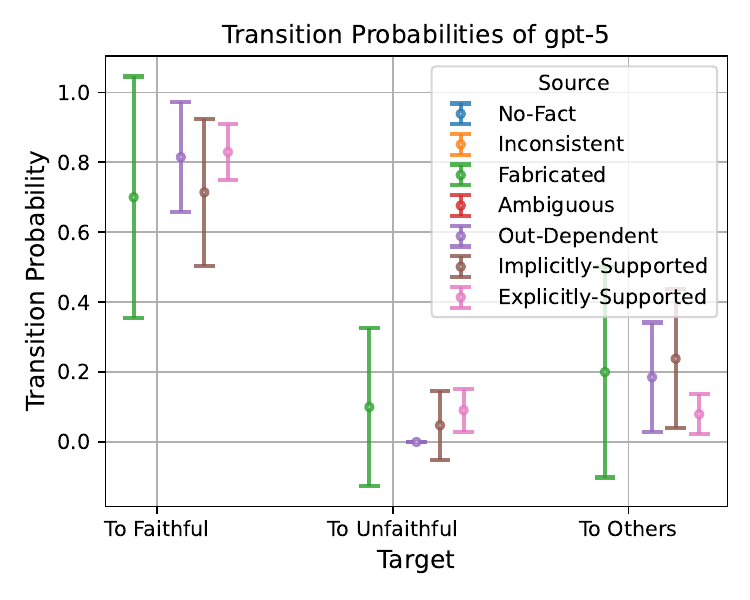}
    }
    \subfigure{
    \includegraphics[width=0.23\textwidth]{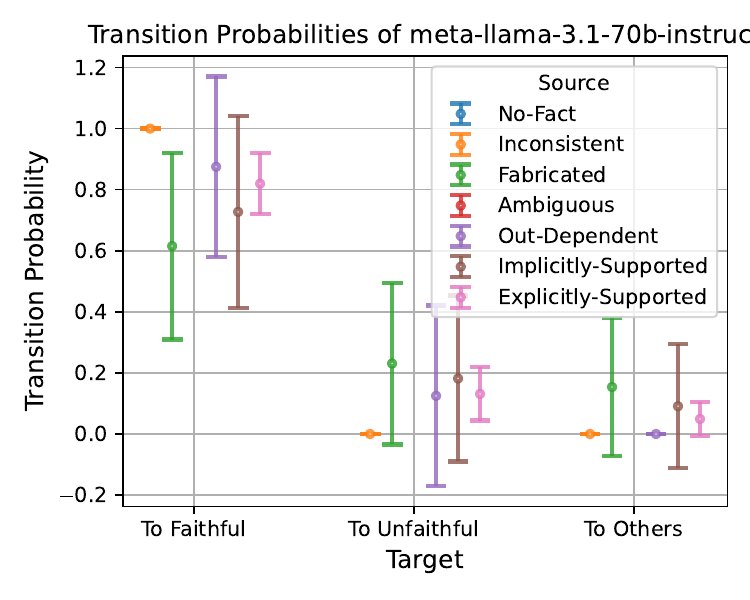}
    }
    \subfigure{
    \includegraphics[width=0.23\textwidth]{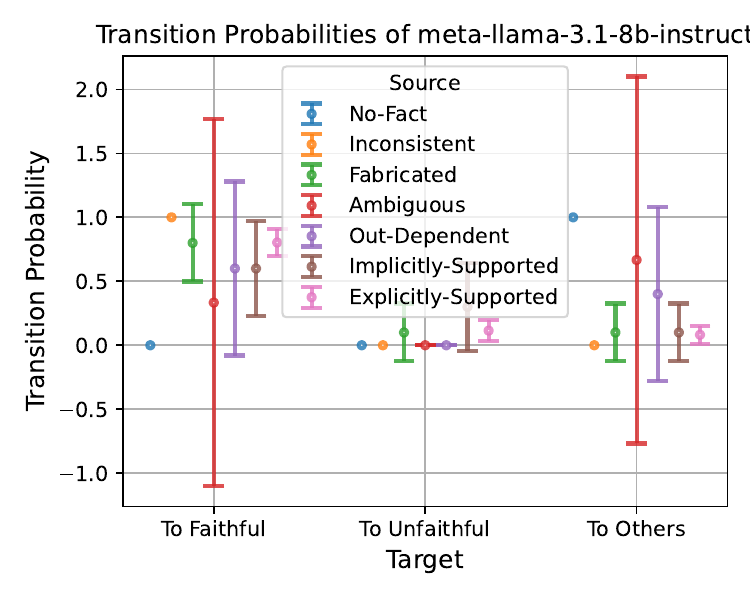}
    }
    \subfigure{
    \includegraphics[width=0.23\textwidth]{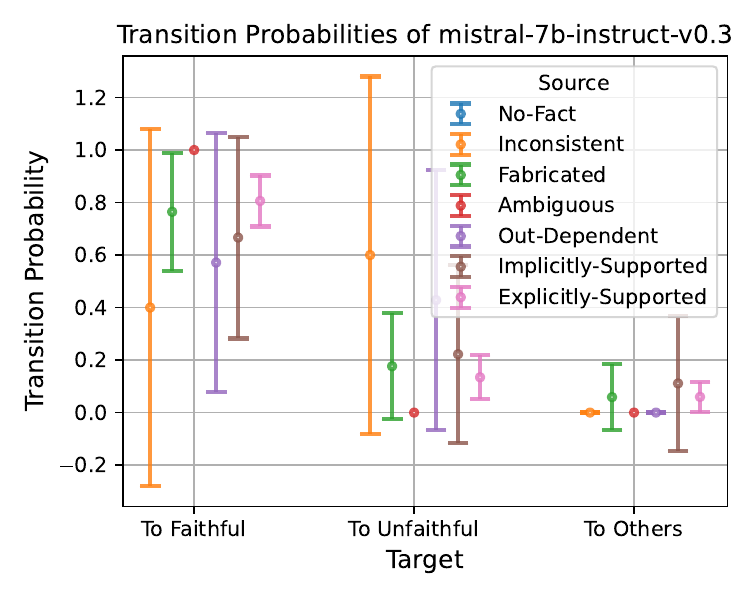}
    }
    \subfigure{
    \includegraphics[width=0.23\textwidth]{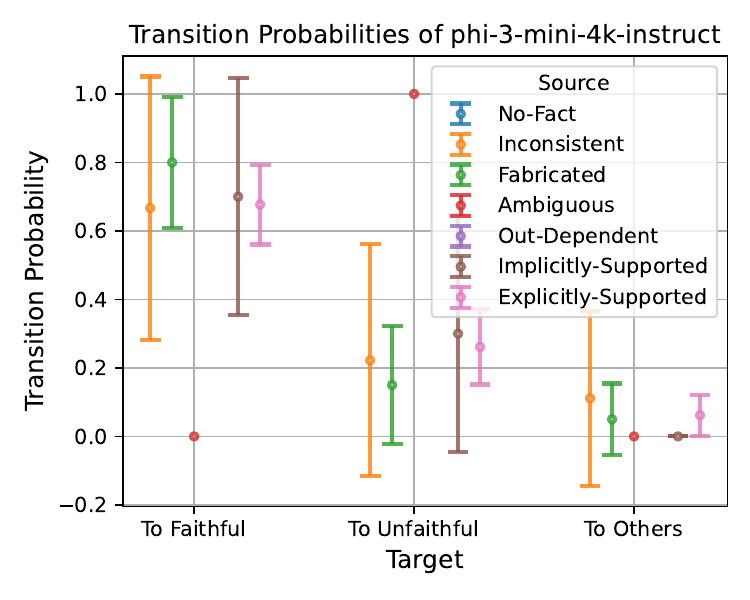}
    }
    \subfigure{
    \includegraphics[width=0.23\textwidth]{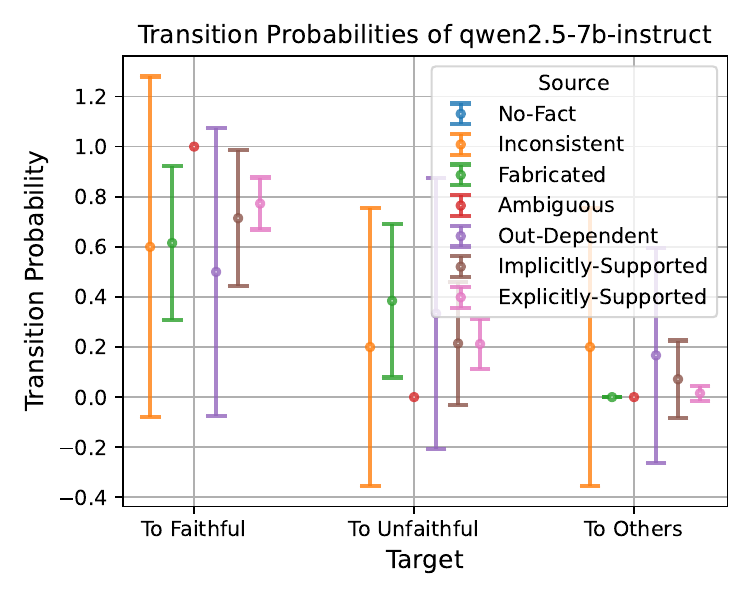}
    }
    \subfigure{
    \includegraphics[width=0.23\textwidth]{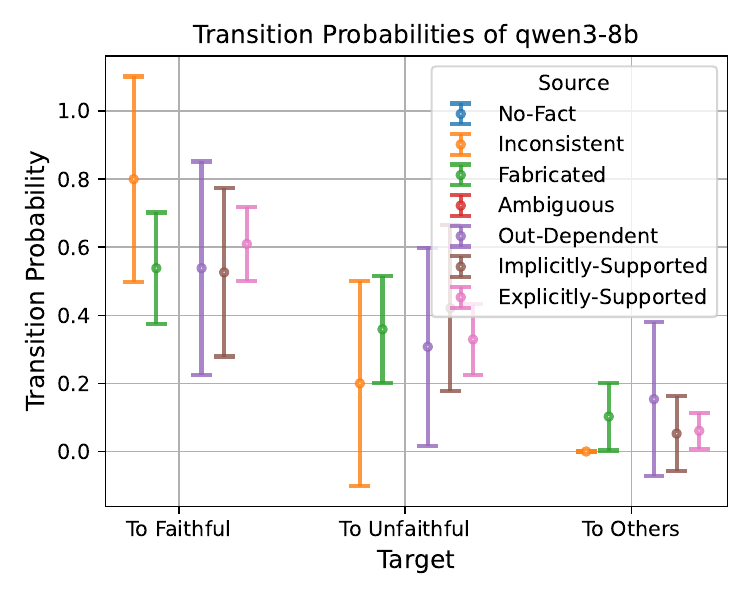}
    }
    \subfigure{
    \includegraphics[width=0.23\textwidth]{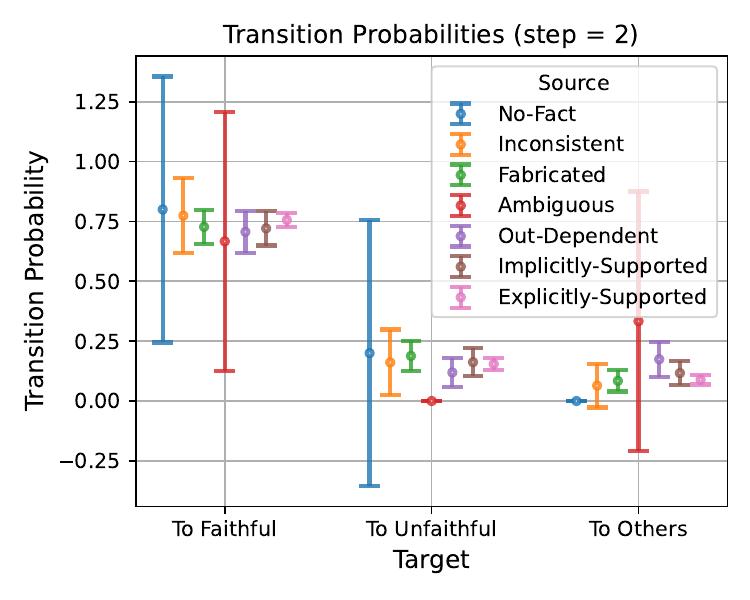}
    }
    \subfigure{
    \includegraphics[width=0.23\textwidth]{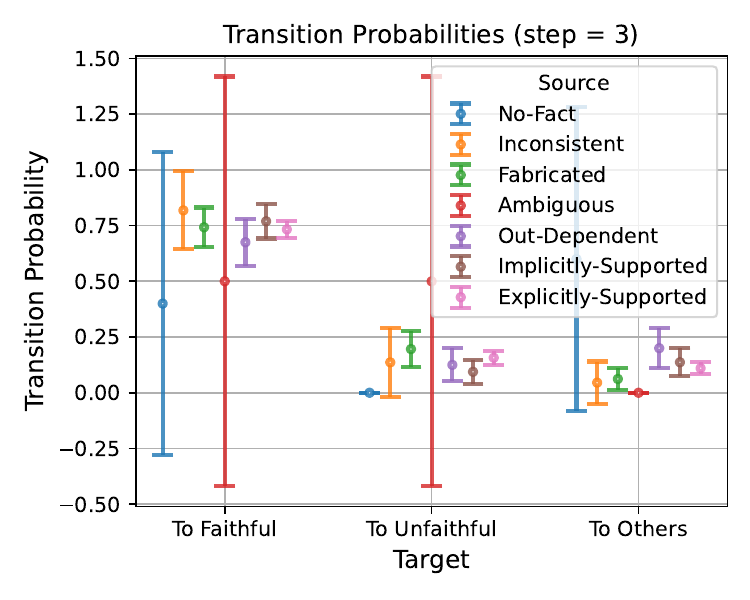}
    }    
    \subfigure{
    \includegraphics[width=0.23\textwidth]{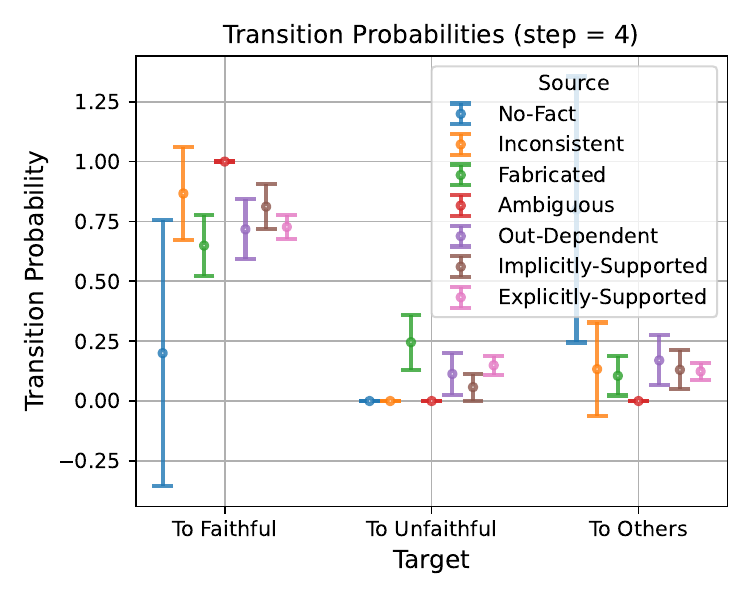}
    }    
    \caption{The model-wise and longer-dependency estimated transition probabilities (with confidence of 95\%).}
    \label{fig:full-transition-probabilities}
\end{figure*}

%\noindent\textbf{Sequential Dependency of Unfaithfulness.} 
The phenomenon of \textit{hallucination snowballing} \citep{zhang2024language} -- where earlier incorrect claims trigger subsequent hallucinated explanations -- has been observed in several multi-step question-answering datasets, indicating a sequential dependency of hallucinations.
To examine whether such sequential dependency exists in our benchmark, we model sentence-wise faithfulness in each response as a Markov chain and estimate the transition probabilities among unfaithfulness classes (see Figure~\ref{fig:transition-probabilities}).
As the results show, the confidence intervals of $\hat p(\text{Faithful} \mid \cdot)$ (and similarly for $\hat p(\text{Unfaithful} \mid \cdot)$) overlap across all source classes (except the outlier of the \texttt{No-Fact} class due to data scarcity).
Comparable results are observed for model-specific and longer-dependency transition probabilities.
The model-wise estimated transition probabilities and the estimated transition probabilities of every 2, 3, 4 steps are shown in Figure~\ref{fig:full-transition-probabilities}.
The results are similar to the overall results in Figure~\ref{fig:transition-probabilities}.
Thus, the sequential dependency of unfaithfulness in our dataset is weak, suggesting that prior sentence hallucinations do not provide a reliable shortcut for predicting unfaithfulness in subsequent sentences.

\section{Implementation Details of Baselines}
\label{sec:baselines-implementation}

The NLI model used in InFusE is the \texttt{cross-encoder/nli-deberta-v3-large}\footnote{\url{https://huggingface.co/cross-encoder/nli-deberta-v3-large}} model on HuggingFace.
The CCP, LLM-Check, and ReDeEP originally output uncertainty scores rather than a 0-1 decision.
We map the score to a 0-1 faithfulness prediction using a threshold $\tau$, with scores greater than $\tau$ mapped to the Unfaithful class.
The CCP used a threshold of -0.001.
The LLM-Check was implemented using a base model of Llama-2-7B \citep{touvron2023llama} and a threshold of -174.0/6.3 for Attn Score / Hidden Score, respectively, where the Attn Score / Hidden Score is extracted from Layer 21/20. 
The ReDeEP was implemented with a base model of Llama-2-7B, $\alpha = 1, \beta = 1.6$, and a threshold of 0.6, where the selected copy heads were the top-7 scoring copy heads, and the selected FFN layers were the top-3 layers, following the settings of ReDeEP (chunk) in \citet{sun2024redeep}.

\section{Zero-shot Detector Details}
\label{sec:zero-shot-details}

\noindent\textbf{Temperature.} All models decode with temperature of 0.6, except GPT-5, whose temperature is not adjustable via API.

\noindent\textbf{Zero-shot Prompt}. The prompt for zero-shot LLM detectors without RAG is as follows, where <DOCUMENT>, <SUMMARY>, and <SENTENCE> are the placeholders of the document, summary, and the sentence, respectively.

\begin{tcolorbox}[title=Detector w/o RAG Prompt, breakable, enhanced jigsaw]
You are judging the faithfulness of the assigned sentence of a summary to the source document. The faithfulness has five options: 

A. Explicitly-Supported: all atomic facts of the sentence appear verbatim (up to a lexical or syntactic transformation) within the document.

B. Generally-Supported: the document entails the sentence, and the sentence is not explicitly-supported by the document. Note that some difference between part of the sentence and the document is allowed, but it should be restricted to the case in which the part of the sentence adopts a weaker or less certain utterance than the document. If any part of the sentence adopts a stronger or more certain utterance than the document, consider option D (Fabricated).

C. Inconsistent: the sentence logically contradicts the document.

D. Fabricated: the sentence does not logically contradict the document, and is neither logically implied by the document nor by external world knowledge.

E. Out-Dependent: the sentence is not logically implied by the document but by the union of the document and external world knowledge.

F. Ambiguous: the sentence or the document has multiple interpretations.

G. No-Fact: the sentence is devoid of facts.

Your task is to decompose the target sentence into atomic facts and then output the correct option.

[Document] <DOCUMENT>

[Summary] <SUMMARY>

[Sentence in the summary] <SENTENCE>

Your option:
\end{tcolorbox}

The prompt for zero-shot LLM detectors enhanced with RAG is as follows, where <DOCUMENT>, <RETRIEVED RESULTS>, <SUMMARY>, and <SENTENCE> are the placeholders of the document, retrieved results, summary, and the sentence, respectively.

\begin{tcolorbox}[title=Detector w/ RAG Prompt, breakable, enhanced jigsaw]    
You are judging whether a given sentence extracted from a summary is faithful to its source document. Several text snippets are retrieved to provide external world knowledge of the source document. The faithfulness has five options: 

A. Explicitly-Supported: all atomic facts of the sentence appear verbatim (up to a lexical or syntactic transformation) within the source document.

B. Generally-Supported: the source document entails the sentence, and the sentence is not explicitly-supported by the document. Note that some difference between part of the sentence and the document is allowed, but it should be restricted to the case in which the part of the sentence adopts a weaker or less certain utterance than the document. If any part of the sentence adopts a stronger or more certain utterance than the document, consider option D (Fabricated).

C. Inconsistent: the sentence logically contradicts the source document.

D. Fabricated: the sentence does not logically contradict the source document, and is neither logically implied by the source document nor by the retrieved snippets.

E. Out-Dependent: the sentence is not logically implied by the source document but by the union of the source document and the retrieved snippets.

F. Ambiguous: the sentence or the source document has multiple interpretations, which could lead to different conclusions.

G. No-Fact: the sentence is devoid of facts.

Your task is to decompose the target sentence into atomic facts and then output the correct option.

[Source Documents] <DOCUMENT>

[Retrieved Snippets] <RETRIEVED RESULTS>

[Summary] <SUMMARY>

[Sentence in the Summary] <SENTENCE>

Your option:
\end{tcolorbox}

\noindent\textbf{RAG Details}. We use the web pages collected for the \texttt{Out-Dependent} class during the annotation process as the retrieval corpus.
The web page texts (in HTML) are first converted into Markdown, and then truncated into snippets of 1600 characters each, with overlaps of 200 characters, resulting in 3030 snippets.
The query for the retrieval is the sentence to verify.
For each query, we retrieved the top 3 documents using the BM25 retriever \citep{robertson1995okapi}.

\section{The Confusion Matrix of GPT-5 + RAG}
\label{sec:confusion}

\begin{figure}[thb]
    \centering
    \includegraphics[width=\linewidth]{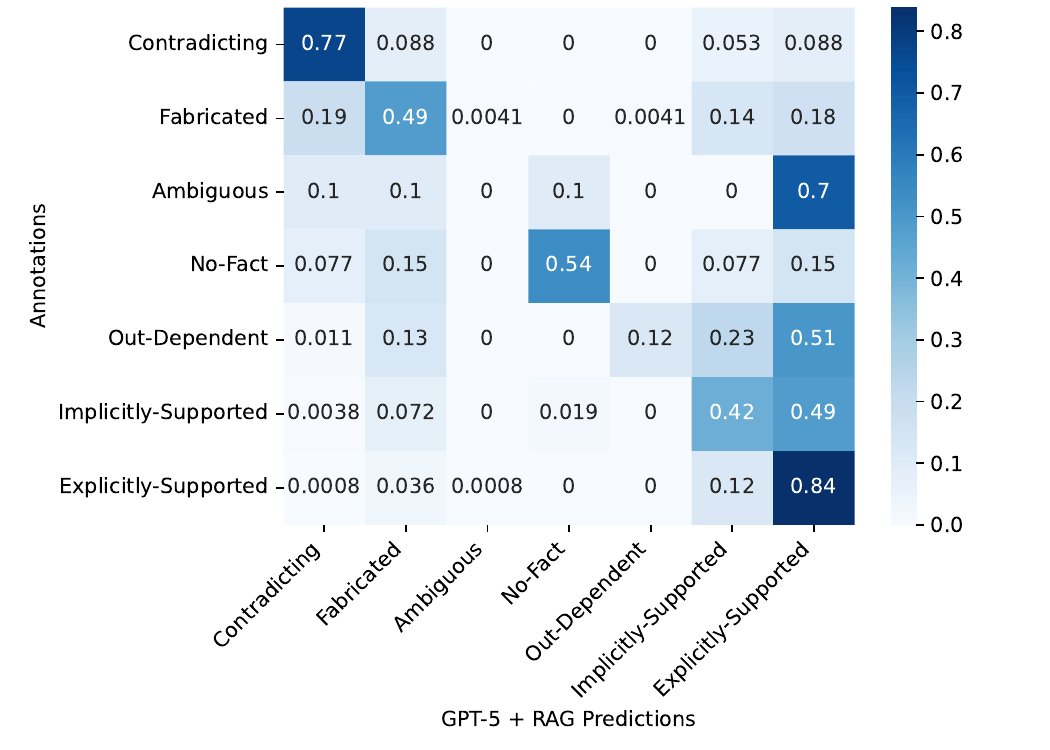}
    \caption{The confusion matrix (with row-wise normalization) of GPT-5 + RAG.}
    \label{fig:gpt-5-rag-confusion}
\end{figure}

The confusion matrix of GPT-5 + RAG is shown in Figure~\ref{fig:gpt-5-rag-confusion}.

\end{document}